\documentclass[journal]{IEEEtran}

\ifCLASSINFOpdf
\else
   \usepackage[dvips]{graphicx}
\fi
\usepackage{url}

\usepackage{graphicx}

\usepackage{multirow}
\usepackage{subfigure}
\usepackage{amsmath}
\usepackage{amssymb}
\usepackage[capitalize]{cleveref}

\usepackage{algorithm}
\usepackage{algpseudocode}

\newtheorem{theorem}{Theorem}
\newtheorem{lemma}{Lemma}
\begin{document}

\title{Neighborhood-Adaptive Generalized Linear Graph Embedding with Latent Pattern Mining}

\author{Shuai Peng, Liangchen Hu, Wensheng Zhang, Biao Jie, and Yonglong Luo
\thanks{This work was supported in part by the Natural Science Project in Universities of Anhui Province under Grant 2022AH040028, and in part by the National Natural Science Foundation of China under Grant 62306010, Grant 62272006, and Grant 61976006, and in part by the Anhui Normal University High-Peak and Incentive Discipline Construction Project under Grant 2023GFXK171. \it{(Corresponding author: Liangchen Hu.)}}
\thanks{Shuai Peng, Liangchen Hu, Biao Jie, and Yonglong Luo are with the School of Computer and Information, Anhui Normal University, Wuhu 241002, China, and also with the Anhui Provincial Key Laboratory of Industrial Intelligent Data Security, Wuhu 241002, China \protect (e-mail: pengshuai@ahnu.edu.cn; cslc.hu@ahnu.edu.cn; jbiao@ahnu.edu.cn; ylluo@ustc.edu.cn).}
\thanks{Wensheng Zhang is with the Institute of Automation, State Key Laboratory of Multimodal Artificial Intelligence Systems, Chinese
Academy of Sciences, Beijing 100190, China, and also with the School of Artificial Intelligence, University of Chinese Academy of Sciences, Beijing 100049, China (e-mail: zhangwenshengia@hotmail.com).}}

\markboth{Journal of \LaTeX\ Class Files, Vol. 14, No. 8, August 2015}
{Shell \MakeLowercase{\textit{et al.}}: Bare Demo of IEEEtran.cls for IEEE Journals}
\maketitle

\begin{abstract}
Graph embedding has been widely applied in areas such as network analysis, social network mining, recommendation systems, and bioinformatics. However, current graph construction methods often require the prior definition of neighborhood size, limiting the effective revelation of potential structural correlations in the data. Additionally, graph embedding methods using linear projection heavily rely on a singular pattern mining approach, resulting in relative weaknesses in adapting to different scenarios. To address these challenges, we propose a novel model, Neighborhood-Adaptive Generalized Linear Graph Embedding (NGLGE), grounded in latent pattern mining. This model introduces an adaptive graph learning method tailored to the neighborhood, effectively revealing intrinsic data correlations. Simultaneously, leveraging a reconstructed low-rank representation and imposing $\ell_{2,0}$ norm constraint on the projection matrix allows for flexible exploration of additional pattern information. Besides, an efficient iterative solving algorithm is derived for the proposed model. Comparative evaluations on datasets from diverse scenarios demonstrate the superior performance of our model compared to state-of-the-art methods.
\end{abstract}


\IEEEpeerreviewmaketitle

\section{Introduction}
\IEEEPARstart{A}{midst} the challenges of processing high-dimensional data in the contemporary landscape, graph embedding technology emerges as a potent tool \cite{Cai20181616,Wang2022415}. It effectively maps intricate graph structures into a low-dimensional vector space, providing a fresh avenue for data analysis. Graph embedding transcends mere dimensionality reduction \cite{LU2022108844,JIANG2023108817}, emphasizing the preservation of the original data's topological structure and relationships within the reduced space. This ensures that the similarity between nodes persists in the new representation. Notably, among these methods, graph embedding approaches that utilize linear projection for out-of-sample extension are particularly intriguing \cite{1388260,1710004,7976386,chen2021adaptive}.

Currently, graph construction methods can be broadly categorized into two main classes: predefined methods \cite{1388260,1710004,7976386,8293687} and dynamically adaptive methods tailored to the data. The latter encompasses four subcategories \cite{9853616}, namely regularization penalty \cite{10.1145/2623330.2623726,wang2018flexible,8608009}, power weighting \cite{10.1145/3369870,9337180}, norm-induced self-weighting \cite{9464126}, and non-negative self-representation \cite{lu2015low,yi2019joint}. 
However, graph embedding methods primarily rely on a single pattern mining approach, typically maintaining only the similarity between nodes in the graph.
Therefore only preserving a kind of structure usually cannot obtain the consistent good performance for different datasets.

In recent years, the low-rank-based feature extraction method has received a lot of attention owning to its robustness to noise and good performance in uncovering the intrinsic global structure of data \cite{Lai2020DiscriminativeLP,9746167,10.1145/3582698,10.1049/cvi2.12228}. 
However, it is a pity that the original low-rank representation (LRR) \cite{6180173} cannot deal with new samples which are not involved in the training stage. 
To address this issue, Latent LRR (LatLRR) takes the hidden data into account and learns a low-rank projection to extract salient features for unsupervised classification \cite{6126422}. 
Double LRR (DLRR) is also proposed for projection learning, in which the principle component recovery and salient feature extraction terms of LatLRR are integrated into a term \cite{6738777}. 
Although the above LRR-based methods have shown their robustness in feature extraction, they lose the ability of dimensionality reduction since the obtained projection has the same dimension with the original data. 
But there are many methods have been proposed to tackle this problem subsequently \cite{7893780,8290566,8827550}.

To integrate the low-rank property, some graph embedding methods were extended for robust feature extraction, such as low-rank preserving projection (LRPP) \cite{lu2015low}, low-rank preserving embedding (LRPE) \cite{ZHANG2017112}, low-rank linear embedding (LRLE) \cite{8356587} and low-rank preserving projection via graph regularized reconstruction (LRPP\_GRR) \cite{8293687}.
the above-mentioned methods tend to utilize multiple pattern in mining intrinsic structure of the data.
However, in this field, there are still some core issues. 
Firstly, current graph construction methods often require the predefined specification of neighborhood size (including adaptive graph learning method), limiting the effective revelation of potential structural relationships in the data. 
This rigid neighborhood definition may fail to capture the diversity and complexity present in real-world data. 
Secondly, the above feature extraction methods do not have strong interpretability since the extracted features are the linear combinations of all the original features. 
Thus it is difficult to interpret which features are the most important for the given task. 
In other words, all original high-dimensional features include outliers and noises are regarded as useful features in these methods. 
This is contradictory with the purpose of the feature extraction which aims to extract useful and discriminative features.

To address these challenges, we will focus on the linear graph embedding and strive to enhance graph construction methods, improve their ability to reveal the latent structure of the data, and conduct in-depth research into more flexible and diverse graph embedding techniques. 
By delving into dynamic adaptive graph construction methods, we aim to enhance the adaptability of graph embedding techniques across diverse scenarios and complex data structures. 
To achieve the interpretability of projection, the $\ell_{2,0}$ norm is introduced to constrain the projection matrix owning to its column-sparsity property. 
In this way, NGLGE can select the most important features from the original data for feature extraction.
This research provides new insights into high-dimensional data dimensionality reduction, driving the further development of graph embedding technologies in practical applications.

In brief, our study has the following key contributions:
\begin{itemize}
  \item We presents a novel unsupervised projection learning approach which utilize low-rank property to preserve global and learn a adaptive weight graph to describe local structures.
  In addition, feature selection also be integrate to our model to filtered redundancy feature.
  \item To solve the proposed model, we developed an efficient iterative algorithm and conducted experiments on various publicly available datasets in different scenarios, demonstrating that our algorithm achieves excellent performance in dimensionality reduction.
  \item Specially, different with previous works, we develop a novel but simple algorithm to optimize the Neighborhood-Adaptive graph which can effectively and exactly find the analytical global optimal solution, then mathematical proofs were subsequently given.
\end{itemize}

The remainder of this paper is structured as follows.
Section \ref{sec:relat:work} provides some notations in our writing and offers a concise overview of related approaches in the field. 
Next, in Section \ref{sec:metho}, we present our novel model, discuss the optimization process, and analyze its computational complexity. 
The experimental setup and results are detailed in Section \ref{sec:exp}.
Finally, we conclude the paper in Section \ref{sec:con}.

\section{Related work}\label{sec:relat:work}
\subsection{Notations and definitions}
Assuming a data matrix $X=[X_1,X_2,...,X_c]=[x_1,x_2,...,x_n]\in \mathbb{R}^{d\times n}$ consisting of $n$ samples in a $d$-dimensional space, where each $X_i$ represents a sub-matrix belonging to the $i$-th class.
For any vector $x=[x_1,x_2,...,x_n]$, the $\ell_2$-norm is defined as $\|x\|_2=\sqrt{\sum^n_{i=1}x_i^2}$ and the $\ell_0$-norm is defined as the number of non-zero entries.
For any matrix $X\in\mathbb{R}^{d\times n}$, the Frobenius norm is defined as $\|X\|_F^2=\sum^n_{j=1}\sum^d_{i=1}x_{ij}^2$, the $\ell_{2,1}$-norm is defined as $\|X\|_{2,1}=\sum^n_{j=1}\sqrt{\sum^d_{i=1}x_{ij}^2}$, the $\ell_{2,0}$-norm is defined as $\|X\|_{2,0}=\|\eta\|_0$, where the $j$-th entry of vector $\eta$ is the $\ell_2$ norm of the $j$-th column vector of $X$. 
$\|X\|_{*}$ represents the nuclear norm of $X$ which is the sum of singular values.

\subsection{Adaptive neighbors learning}
To overcome the disadvantage of the similarity measurement and projection learning are often conducted in two separated steps, Nie et al. \cite{10.1145/2623330.2623726} proposed a adaptive neighbors learning model.
The goal is to find an optimal subspace on which the adaptive neighboring is performed.
Denote the total scatter matrix by $S_t=XHX^T$, where $H$ is the centering matrix defined by $H=I-\frac{1}{n}\mathbf{1}\mathbf{1}^T$ and then, the model is as follows:
\begin{equation}
\begin{aligned}
&\min_{Q,S}\sum^n_{i=1}\sum^n_{j=1}(\|Qx_i-Qx_j\|^2_2s_{ij}+\gamma s_{ij}^2) \\
&~~\mathrm{s.t.}~~QS_tQ^T=I,~S\geq0,~S\mathbf{1}=\mathbf{1}
\end{aligned}
\end{equation}
where $QS_tQ^T=I$ can result in a statistically uncorrelated subspace.
Note that the parameter $\gamma$ is not predetermined but determined during the optimization process by specifying the number of neighbors.

\subsection{Efficient sparse feature selection}
In \cite{8386668}, Pang et al. proposed a novel algorithm which can solve the original $\ell_{2,0}$-norm constrained sparse based feature selection problem directly instead of its approximate problem.
their optimization function can be formulated as follows:
\begin{equation}
\begin{aligned}
&\min_{Q,\mathbf{b}}\|QX+\mathbf{1b}-Y\|_F^2 \\
&~~\mathrm{s.t.}~~\|Q\|_{2,0}=\alpha
\end{aligned}
\label{ESFS}
\end{equation}
where $\mathbf{b}$ is bias and $Y$ is regression label.
The $\ell_{2,0}$-norm constraint of $Q$ makes the number of its non-zero columns be equal to $\alpha$, which ensures the sparsity of the model and makes the model have the ability to select features.
In this article, we use the same strategy to optimize our model.

\section{Methodology}\label{sec:metho}
In this section, we present a method named NGLGE to learn the optimal projection for unsupervised classification.
To solve (\ref{model}) effectively, we develop an efficient ADMM-based optimization algorithm. 
Additionally, analysis of the computational complexity involved in the algorithm is also presented.

\subsection{Proposed method}
Linearly projecting graph nodes into a low-dimensional space while minimizing dissimilarity between pairs of nodes in the embedded space to preserve the local neighborhood of nodes in the original space. The objective is defined as follows
\begin{equation}\label{obj:orginal}
\min_{Q}\sum_{i=1}^n\sum_{j=1}^n\|Qx_i-Qx_j\|_2^2s_{ij}
\end{equation}
where the projection matrix $Q\in\mathbb{R}^{m\times d}(m<d)$ transforms $x_i$ into a lower-dimensional space, and $s_{ij}$ denotes the affinity between $x_i$ and $x_j$. In model (\ref{obj:orginal}), relying solely on the preservation of the adjacency of nodes in the graph to ensure the exploration of latent patterns in the data is overly simplistic.

In \cite{6738777}, discriminative structural information is adeptly extracted from observed data through the concurrent recovery of information from both row and column spaces, as follows
\begin{equation}\label{DLRR}
\min_{Z,L,E}\|Z\|_{*}+\|L\|_{*}+\lambda\|E\|_1~~~~\mathrm{s.t.}~~X=LXZ+E
\end{equation}
where $L\!\in\!\mathbb{R}^{d\times d}$ and $Z\!\in\!\mathbb{R}^{n\times n}$ are two low-rank representation matrices, $E$ contains noise or errors, and $\lambda$ is a regularization parameter. To enrich the disclosure of latent patterns in the data, we hereby examine the correlation between the original $X$ and the intrinsic $LXZ$ below
\begin{equation}\label{obj:improved}
\min_{Z,L}\sum_{i=1}^n\sum_{j=1}^n\|x_i-LXz_j\|_2^2s_{ij}+\lambda_1\|L\|_{*}+\lambda_2\|Z\|_{*}
\end{equation}
where $\lambda_1$ and $\lambda_2$ are two regularization parameters. Considering $L$ as a low-rank matrix, it can be expressed as $L=PQ$, where $P\in\mathbb{R}^{d\times m}$, indicating that $\mathrm{rank}(L)\!\leq\!m\!<\!d$. With the constraint $P^TP = I$, (\ref{obj:improved}) can be expressed as
\begin{equation}\label{fixedgraph}
\begin{aligned}
&\min_{P,Q,Z}\sum^n_{i=1}\sum^n_{j=1}\|x_i\!-\!PQXz_j\|^2_2s_{ij}+\lambda_1\|Q\|_F^2+\lambda_2\|Z\|_{*} \\
&~~\mathrm{s.t.}~~P^TP=I,~\|Q\|_{2,0}=\alpha
\end{aligned}
\end{equation}
where $\|Q\|_{2,0}=\alpha$ compels the number of non-zero columns in the projection matrix $Q$ to be equal to $\alpha$, endowing the model with feature selection capability and enhancing the exploration of latent patterns within the data.

Moreover, in the presence of noise-contaminated data, the pre-constructed graph fails to precisely capture the affinity within the data. In light of this, we make $S$ adaptively to learn probabilistic neighborhood graph, as detailed bellow
\begin{equation}\label{model}
\begin{aligned}
&\min_{P,Q,Z,S}\sum^n_{i=1}\sum^n_{j=1}\|x_i-PQXz_j\|^2_2s_{ij}+\lambda_1\|Q\|_F^2 \\
&~~~~~~~~~~~~+\lambda_2\|Z\|_{*}+\lambda_3\|S\|_F^2 \\
&~~\mathrm{s.t.}~~P^TP=I,~S\geq0,~S^T\mathbf{1}=\mathbf{1},~\|Q\|_{2,0}=\alpha
\end{aligned}
\end{equation}

\subsection{Optimization}
In this section, we address the problem presented in (\ref{model}) using the alternating direction method of multipliers (ADMM) framework \cite{MAL-016}. 
A detailed convergence analysis of ADMM for nonconvex problems is available in \cite{doi:10.1137/140990309}. 
To facilitate the optimization process, we introduce an auxiliary variable $B$ and impose a constraint $Z\!=\!B$ to achieve separability in (\ref{model}). 
Thus, by relaxing the original problem, we can reformulate (\ref{model}) as follows
\begin{equation}
\begin{aligned}
&\min_{P,Q,Z,S}\sum^n_{i=1}\sum^n_{j=1}\|x_i-PQXz_j\|^2_2s_{ij}+\lambda_1\|Q\|_F^2 \\
&~~~~~~~~~~~~+\lambda_2\|B\|_{*}+\lambda_3\|S\|_F^2 \\
&~\mathrm{s.t.}~~P^TP=I,~S\geq0,~S^T\mathbf{1}=\mathbf{1},~\|Q\|_{2,0}=\alpha,~Z=B
\end{aligned}
\end{equation}

For convenience of taking derivative, we rewrite the first and second terms in trace form.
\begin{equation}
\begin{aligned}
&\min_{P,Q,Z,S}\mathrm{Tr}(X^TXD_1)-2\mathrm{Tr}(X^TPQXZS^T) \\
&~~~~~~~~~~~~+\mathrm{Tr}(Z^TX^TQ^TQXZD_2)+\lambda_1\mathrm{Tr}(Q^TQ) \\
&~~~~~~~~~~~~+\lambda_2\|B\|_{*}+\lambda_3\|S\|_F^2 \\
&~\mathrm{s.t.}~~P^TP=I,~S\geq0,~S^T\mathbf{1}=\mathbf{1},~\|Q\|_{2,0}=\alpha,~Z=B
\end{aligned}
\label{modeltrace}
\end{equation}
where $D_1=\mathrm{diag}(\sum^n_{j=1}s_{1j},\sum^n_{j=1}s_{2j},\dots,\sum^n_{j=1}s_{nj})$ and $D_2=\mathrm{diag}(\sum^n_{i=1}s_{i1},\sum^n_{i=1}s_{i2},\dots,\sum^n_{i=1}s_{in})$.

Then, the augmented Lagrangian function of (\ref{modeltrace}) is defined by
\begin{equation}
\begin{aligned}
&\mathcal{L}(P,Q,Z,S,B) \\
&~=\mathrm{Tr}(X^TXD_1)-2\mathrm{Tr}(X^TPQXZS^T) \\
&~~~~~+\mathrm{Tr}(Z^TX^TQ^TQXZD_2)+\lambda_1\mathrm{Tr}(Q^TQ) \\
&~~~~~+\lambda_2\|B\|_{*}+\lambda_3\|S\|_F^2+\left\langle C,Z-B\right\rangle \\
&~~~~~+\frac{\mu}{2}\|Z-B\|_F^2 \\
&~\mathrm{s.t.}~~P^TP=I,~S\geq0,~S^T\mathbf{1}=\mathbf{1},~\|Q\|_{2,0}=\alpha
\end{aligned}
\end{equation}
where $\left\langle A,B\right\rangle=\mathrm{trace}(A^TB)$.
Note that $D_2$ is a identity matrix, so we ignore it later.

\textbf{Update} \boldmath$Z$\unboldmath\textbf{:} Given fixed $P$, $Q$, $S$ and $B$, the subproblem for solving $Z$ is as follows
\begin{eqnarray}
\begin{aligned}
&\min_{Z}\mathrm{Tr}(Z^TX^TQ^TQXZ)-2\mathrm{Tr}(X^TPQXZS^T) \\
&~~~~~~~+\left\langle C,Z-B\right\rangle+\frac{\mu}{2}\|Z-B\|_F^2
\end{aligned}
\end{eqnarray}
Let ${\partial\mathcal{L}(Z)}/{\partial Z}=0$, we have
\begin{equation}
(2X^TQ^TQX+\mu I)Z=2X^TQ^TP^TXS+\mu B-C
\end{equation}
Then $Z$ is updated by
\begin{equation}
Z=(2X^TQ^TQX+\mu I)^{-1}(2X^TQ^TP^TXS+\mu B-C) \label{updateZ}
\end{equation}

\textbf{Update} \boldmath$B$\unboldmath\textbf{:} Given fixed $Z$, the subproblem concerning $B$ is as follows
\begin{equation}
\min_{B}\lambda_2\|B\|_{*}+\frac{\mu}{2}\|Z-B+\frac{C}{\mu}\|_F^2
\end{equation}
Then $B$ is obtained by using the singular value thresholding
(SVT) shrinkage operator \cite{8827550} as follows
\begin{equation}
B=\Theta_{\lambda_2/\mu}\left(Z+\frac{C}{\mu}\right) \label{updateB}
\end{equation}
where $\Theta$ is the SVT shrinkage operator.

\textbf{Update} \boldmath$Q$\unboldmath\textbf{:} Given fixed $P$, $Z$ and $S$, the problem w.r.t. $Q$ becomes
\begin{eqnarray}
\begin{aligned}
&\min_{Q}\mathrm{Tr}(Z^TX^TQ^TQXZ)-2\mathrm{Tr}(X^TPQXZS^T) \\
&~~~~~~~+\lambda_1\mathrm{Tr}(Q^TQ) \\
&~~\mathrm{s.t.}~~\|Q\|_{2,0}=\alpha
\end{aligned}
\label{Q}
\end{eqnarray}
where $m\leq\alpha\leq d$.
Motivated by \cite{8386668}, we conduct full rank decomposition on $Q$, i.e.,
\begin{equation}
Q=VU \label{Q=VU}
\end{equation}
where $V\in \mathbb{R}^{m\times\alpha}$ is composed by all non-zero columns of $Q$ and $U\in \mathbb{R}^{\alpha\times d}$ is a selection matrix.
It is worth noticing that the full rank decomposition is not unique.
Here, for the purpose of feature selection, we impose constraints on $U$
\begin{equation}
U=[\mathbf{u}_{s(1)};\mathbf{u}_{s(2)};\dots;\mathbf{u}_{s(\alpha)}]
\end{equation}
where the vector $\mathbf{s}$ is a permutation of $\{1,2,\dots,d\}$ and $\mathbf{u}_i$ is the $i$-th row of identity matrix $I\in\mathbb{R}^{d\times d}$.
By replacing $Q$ in problem (\ref{Q}) with (\ref{Q=VU}), we have
\begin{eqnarray}
\begin{aligned}
&\min_{U\in selec,V}\mathrm{Tr}[VU(XZZ^TX^T+\lambda_1I)U^TV^T] \\
&~~~~~~~~~~~~~~~~-2\mathrm{Tr}(VUXZS^TX^TP)
\end{aligned}
\label{minUV}
\end{eqnarray}
Then, after taking derivative of problem (\ref{minUV}) w.r.t. $V$ and assigning this value to zero, we get
\begin{equation}
V=P^TXSZ^TX^TU^T[U(XZZ^TX^T+\lambda_1I)U^T]^{-1}
\label{V}
\end{equation}
Then, substituting (\ref{V}) into (\ref{minUV}) and denoting $F=P^TXSZ^TX^T$, $G=XZZ^TX^T+\lambda_1I$, we have
\begin{eqnarray}
\begin{aligned}
&\min_{U\in selec}\mathrm{Tr}[FU^T(UGU^T)^{-1}UF^T] \\
&~~~~~~~~~~~~~~~-2\mathrm{Tr}[FU^T(UGU^T)^{-1}UF^T]
\end{aligned}
\label{tr-2tr}
\end{eqnarray}
Therefore, problem (\ref{tr-2tr}) becomes
\begin{equation}
\max_{U\in selec}\mathrm{Tr}[(UGU^T)^{-1}UF^TFU^T]
\label{maxU}
\end{equation}
To maximize problem (\ref{maxU}), we select the largest $\alpha$ diagonal elements of $(G^{-1})(F^TF)$ and denote the row (or column) indices vector of the largest $\alpha$ diagonal elements as $\mathbf{q}=\{q_1,q_2,\dots,q_\alpha\}$, where $q_\alpha\in \{1,2,\dots,d\}$.
Then, $U_{iq_i}=1, i=1,2,\dots,\alpha$ while other element is zero \cite{ZHU202368}.

\textbf{Update} \boldmath$P$\unboldmath\textbf{:} Given fixed $Q$, $Z$ and $S$, update $P$ by solving the following maxmization problem:
\begin{eqnarray}
\begin{aligned}
&\max_{P}\mathrm{Tr}(X^TPQXZS^T) \\
&~\mathrm{s.t.}~~P^TP=I
\end{aligned}
\label{opp}
\end{eqnarray}
The problem (\ref{opp}) is a orthogonal procrustes problem (OPP) problem \cite{Zou2006sparse} which can be simply solved by singular value decomposition (SVD).
We compute the thin SVD of $XSZ^TX^TQ^T$, i.e.,
\begin{equation}
XSZ^TX^TQ^T=U_{d\times m}\Sigma V^T_{m\times m} 
\end{equation}
where $m\leq d$, then, we get
\begin{equation}
P=UV^T \label{updateP}
\end{equation}
where $U$ and $V$ consist of the left-singular vectors and right-singular vectors of $XSZ^TX^TQ^T$, respectively.

\textbf{Update} \boldmath$S$\unboldmath\textbf{:} Given fixed $P$, $Q$ and $Z$, the subproblem for updating $S$ is
\begin{equation}
\begin{aligned}
&\min_{S}\sum^n_{i=1}\sum^n_{j=1}a_{ij}s_{ij}+\lambda_3\|S\|_F^2 \\
&~\mathrm{s.t.}~~S\geq0,~S^T\mathbf{1}=\mathbf{1}
\end{aligned}
\label{S}
\end{equation}
where $a_{ij}=\|x_i-PQXz_j\|^2_2$.
We break down the problem (\ref{S}) in $n$ independent sub-problems
\begin{equation}
\begin{aligned}
&\min_{\mathbf{s}_j}\sum^n_{i=1}a_{ij}s_{ij}+\lambda_3\|\mathbf{s}_j\|^2_{2} \\
&~\mathrm{s.t.}~~\mathbf{s}_j\geq0,~\sum^n_{i=1}s_{ij}=1
\end{aligned}
\label{subS}
\end{equation}
To optimize this problem, Nie et al. \cite{10.1145/2623330.2623726} proposed a method which has since been frequently adopted.
But in that way, the all training sample have same number of neighbors, which may lead to incorrect connections between samples.
To fix this problem, we develop a new algorithm which can effectively and exactly find the analytical global optimal solution.
We discuss the new algorithm in more detail later and give mathematical proof subsequently.

\textbf{Update} \boldmath$C$ \textbf{and} $\mu$\unboldmath\textbf{:} $C$ and $\mu $ are respectively updated by
\begin{equation}
\begin{array}{l}
C=C+\mu(Z-B) \\
\mu=\min(\rho\mu,~\mu_{max})
\label{C_and_mu}
\end{array}
\end{equation}

The detailed optimization procedure of (\ref{model}) is presented in Algorithm \ref{alg:framwork}.
\begin{algorithm}
\caption{Solving problem (\ref{model}) via ADMM.}
\label{alg:framwork}
\begin{algorithmic}[1]
\Require Training samples $X\in \mathbb{R}^{d\times n}$, hyper-parameters $\lambda_1$, $\lambda_2$, $\lambda_3$ and projected dimension $m$.
\Ensure Discriminative projection matrix $Q$. \\
\textbf{Initialize} $P=\arg\max_{P^TP=I}Tr(P^T\Sigma P)$, where $\Sigma$ is the data covariance; $Q=P^T$, $Z=B=0$, $s_{ij}=1/k$ if $x_i\in N_k(x_j)$, $s_{ij}=0$ otherwise; $C=0$, $\mu=0.1$, $\rho=1.1$, $\mu_{max}=10^8$, $\epsilon=10^{-6}$ and $maxIter=60$.
\Repeat
\State Update $Z$ using (\ref{updateZ}).
\State Update $B$ using (\ref{updateB}).
\State Update $Q$ using (\ref{Q=VU}).
\State Update $P$ using (\ref{updateP}).
\State Update $S$ using (\ref{sopt}).
\State Update $C$ and $\mu$ using (\ref{C_and_mu}).
\Until{the following convergence conditions are fulfilled
\begin{eqnarray}
\|Z-B\|_{\infty}\leq\epsilon\nonumber
\end{eqnarray}
or iteration reaches $maxIter$.
}
\end{algorithmic}
\end{algorithm}

\subsection{Optimization algorithm for updating $S$}
For simple of notation, we discuss the following generalized problem:
\begin{equation}
\begin{aligned}
&\min_{\mathbf{s}}\sum^n_{i=1}a_{i}s_{i}+\gamma\|\mathbf{s}\|^2_{2} \\
&~~\mathrm{s.t.}~~\mathbf{s}\geq0,~\sum^n_{i=1}s_{i}=1
\end{aligned}
\label{Sgeneral}
\end{equation}
where $\mathbf{a}$ and $\mathbf{s}$ are vectors with $n$ elements.
For convenience to characterize the algorithm of this problem, we denote
\begin{align}
S^t&=\{i\mid a_i<c^{t-1}\},~~~~t=1,2,\dots \\
c^t&=\frac{\sum_{i\in S^t}a_{i}}{k_t}+\frac{2\gamma}{k_t},~~~~t=0,1,\dots
\end{align}
where $S^t$ is a index set, $k_t$ is number of elements in $S^t$.
We first let $S^0=\{1,2,\dots,n\}$, then all variables will be defined recursively.
Then, we only need to find the $t$ that makes $c^t=c^{t-1}$ (or $k^t=k^{t-1}$).



In the worst case, it is only necessary to compute up to $c^n$ for us to find $c^t=c^{t-1}$.
If the $t$ has been found, then
\begin{eqnarray}
s_i = \left\{
\begin{array}{ll}
\frac{\sum_{i\in S^t}a_{i}}{2\gamma k_t}-\frac{a_i}{2\gamma}+\frac{1}{k_t}, & i\in S^t \\
0, & \mathrm{otherwise} \\
\end{array}
\right.
\label{sopt}
\end{eqnarray}
Next, we discuss why $\mathbf{s}$ obtained by (\ref{sopt}) is optimal.
\begin{lemma}
The sequence $c^t,~t=0,1,\dots$, is nonincreasing.
\end{lemma}
\begin{IEEEproof}
We only prove that $c^0\geq c^1$ here, it is similar to prove $c^t\geq c^{t+1}$ for any $t$.
Supposing that $a_{1},a_{2},\dots,a_{k},a_{k+1},\dots,a_{n}$ is sorted in ascending order, then, $c^0=\frac{\sum^n_{i=1}a_{i}}{n}+\frac{2\gamma}{n}$.
Without loss of generality, assuming that $a_{1},\dots,a_{k}<c^0$ and $a_{k+1},\dots,a_{n}\geq c^0$.
Then, $c^1=\frac{\sum^k_{i=1}a_{i}}{k}+\frac{2\gamma}{k}$.
We need to prove that $c^0\geq c^1$, e.g.:
\begin{align}
\frac{\sum^n_{i=1}a_{i}}{n} + \frac{2\gamma}{n} \geq \frac{\sum^k_{i=1}a_{i}}{k} + \frac{2\gamma}{k}
\label{77}
\end{align}
The inequality is not obvious, because from left to right, the value of the first term decreases while the value of the second term increases.
But we can prove this inequality from known conditions.
From $a_{k+1},\dots,a_{n}>c^0$, we have:
\begin{align}
\frac{\sum^n_{i=k+1}a_{i}}{n-k} \geq \frac{\sum^n_{i=1}a_{i}}{n} + \frac{2\gamma}{n} \label{abargeqc0}
\end{align}
The left side of (\ref{abargeqc0}) is the mean value of $a_{k+1},\dots,a_{n}$.
Using some simple algebra, yields:
\begin{small}
\begin{align}
&n\sum^n_{i=k+1}a_{i} \geq (n-k)(\sum^n_{i=1}a_{i} + 2\gamma) \\
&n\sum^k_{i=1}\!a_{i}\!+\!n\sum^n_{i\!=\!k\!+\!1}\!a_{i}\!\geq\!n(\sum^n_{i=1}\!a_{i}\!+\!2\gamma)\!-\!k(\sum^n_{i=1}\!a_{i}\!+\!2\gamma)\!+\!n\sum^k_{i=1}\!a_{i} \\
&k\sum^n_{i=1}a_{i} + 2k\gamma \geq n\sum^k_{i=1}a_{i} + 2n\gamma \\
&\frac{\sum^n_{i=1}a_{i}}{n}+\frac{2\gamma}{n}\geq\frac{\sum^k_{i=1}a_{i}}{k}+\frac{2\gamma}{k}\label{66}
\end{align}
\end{small}
where (\ref{66}) is exactly identical to (\ref{77}) which is required to be proven.
\end{IEEEproof}

\begin{theorem}
Computing $\mathbf{s}$ by (\ref{sopt}) yields the optimal solution for problem (\ref{Sgeneral}).
\end{theorem}
\begin{IEEEproof}
The KKT conditions for problem (\ref{Sgeneral}) are as follows
\begin{equation}
\begin{array}{rr}
s_{i} \geq 0,& i = 1,\dots,n \\
\sum^n_{i=1}s_{i} = 1,& \\
\lambda_{i} \geq 0,& i = 1,\dots,n \\
\lambda_{i}s_{i} = 0,& i = 1,\dots,n \\
\mathbf{a} + 2\gamma\mathbf{s} - \lambda + \nu\mathbf{1} = 0,&
\end{array}
\end{equation}
where $\nu$ is a scalar and $\lambda, \mathbf{1}\in\mathbb{R}^n$.
Because problem (\ref{Sgeneral}) is convex problem and the objective function is strongly convex, if we can find $\mathbf{s}$ together with 
($\lambda, \nu$) that satisfy the KKT conditions, $\mathbf{s}$ is optimal \cite{boyd2004convex}.

Because $a_i<c^t$, by (\ref{sopt}), the first and second KKT conditions can be satisfied.
Then calculate $\nu$.
For those $s_{i}>0$, $\lambda_{i}$ is 0.
By the last KKT condition, we have:
\begin{align}
2\gamma s_{i} &= -a_{i} - \nu
\end{align}
Sum both sides of the equation for nonzero $s_{i}$ and solve $\nu$
\begin{align}
\nu &= -\frac{\sum_{i\in S^t}a_{i}}{k_t} - \frac{2\gamma}{k_t}
\end{align}
Finally, we need to find appropriate $\lambda_{i}$ corresponding to zero $s_{i}$.
Those $\lambda_{i}$ must satisfy the third and last KKT conditions:
\begin{align}
\lambda_{i} &\geq 0 \label{lambdageq0}\\
a_{i}-\lambda_{i} &=\frac{\sum_{i\in S^t}a_{i}}{k_t}+\frac{2\gamma}{k_t} \label{lastkktcond}
\end{align}
Note that the right side of (\ref{lastkktcond}) is $c^t$ while $c^t\leq c^{t-1}\leq c^{t-2}\leq\dots\leq c^0$.
Recall that all $a_{i}$ corresponding to zero $s_{i}$ greater than or equal to one of $\{c^0,c^1,\dots,c^t\}$, thus greater than or equal to the $c^t$.

Overall, we can find $\lambda_{i}$ that satisfies both (\ref{lambdageq0}) and (\ref{lastkktcond}).
So $\mathbf{s}$ is optimal because there are ($\lambda, \nu$) that, together with $\mathbf{s}$, satisfy the KKT conditions.
\end{IEEEproof}

\subsection{Complexity analyses}
The computational complexity of the model mainly depends on the complexity of Algorithm \ref{alg:framwork} in which the main time-consuming operations are solving the inverse of square matrix and performing SVD.

Next, we discuss the main computational complexity of each step.
To update $Z$, the main time consuming step is to compute $(2X^TQ^TQX+\mu I)^{-1}$, where the computational complexity is about $O(n^3)$.
To update $B$, the most demanding computation is the SVD computation of matrix $(Z+C/\mu)$, and thus the computational complexity is $O(n^3)$.
To update $Q$, the main time consuming step is to compute $(UGU^T)^{-1}$, in which the computational complexity is about $O(d^3)$.
When updating $P$, the computational complexity is dominated by the singular value decomposition (SVD) computation of $XSZ^TX^TQ^T$, which has a complexity of $O(m^3)$ \cite{8827550}.
The optimization of the $S$ is relatively simple, so the computational cost of solving $S$ can be ignored.
Overall, Algorithm \ref{alg:framwork} has a total computational complexity of $O(\tau(n^3+d^3+m^3))$, where $\tau$ denotes the number of iterations.
In practical scenarios, it is common to have $d\ll n$ and a small value of $m$.
Hence, the total computational complexity of the model can be loosely approximated as $O(\tau n^3)$.

\section{Experiments}\label{sec:exp}
In this section, we evaluate the effectiveness of the proposed method on six widely-used public datasets collected from diverse scenarios.
To ensure a convincing comparison, the proposed method is compared with some state-of-the-art methods including LPP \cite{1388260}, NPE \cite{1544858}, OLPP \cite{1710004}, PCAN \cite{10.1145/2623330.2623726}, SOGFS \cite{10.5555/3015812.3016004}, RJSE \cite{LAI201830}, RDR \cite{8063375}, LRLE \cite{8356587}, LRPP\_GRR \cite{8293687}, FSP \cite{8693535}, LRAGE \cite{LU2021107758}.

\subsection{Databases}

Here, six distinctive and widely-used databases are employed to demonstrate the validity of our algorithm.

\textbf{EYaleB} \cite{927464} - The Extended Yale B database consists of a total of 2414 face images captured from 38 volunteers.
Each volunteer provided 59-64 images with different illuminations.
Moreover, half of the images per person contain varying degrees of shadows.
In the experiments, each sample was resized to 32 $\times$ 32 pixels.

\textbf{YTC} \cite{5995566} - The YouTube-Celebrities (YTC) dataset consists of 1910 video clips featuring 47 celebrities sourced from YouTube.
These videos are obtained from real-life scenarios, exhibiting significant variations in facial expressions, appearances and poses, as well as containing noise, misalignments, poor quality, and occlusions.
Each of the 1910 video clips contains 8-400 frames, and each celebrity is represented by more than 1000 images.
For our experiments, 200 images of each celebrity are randomly selected as the experimental data (i.e., 9400 images in total).
The face images are uniformly resized to 30 $\times$ 30 pixels.

\textbf{Binalpha}\footnote{\url{https://cs.nyu.edu/~roweis/data.html}} - This database consists of 1404 samples distributed among 36 classes.
Each sample is represented by a binary image of 20 $\times$ 16 pixels.
The database encompasses not only digits ranging from ``0" to ``9" but also capital letters from ``A" to ``Z".
Notably, There are several pairs of easily confused letters and numbers such as ``0" and ``O", ``2" and ``Z", ``5" and ``S", ``6" and ``G", ``8" and ``B", thus posing a challenge to classification.

\textbf{USPS} \cite{291440} - The USPS dataset is a collection of handwritten digital images comprising 10 classes from ``0" to ``9".
The dataset contains a total of 9289 images, with each class consisting of varying numbers of samples ranging from 708 to 1553 images.
To facilitate analysis, all images have been resized to 16 $\times$ 16 pixel gray images, so the original dimensional of samples is 256.

\textbf{ETH80} \cite{1211497} - The ETH80 dataset comprises visual object images belonging to 8 different categories including apples, cars, cows, cups, dogs, horses, pears and tomatoes.
For each category, there are 10 object instances, and images of each object instance were captured from 41 different viewpoints.
In our experiment, we resized the gray images to 20 $\times$ 20 pixels and converted them to 400-dimensional vectors.

\textbf{15-Scene} \cite{1641019} -  The 15-Scene database totally includes 4485 images from 15 natural scene categories, which is commonly used in the evaluation of scene recognition.
A variety of scenes can be found in the database, including Bedroom, Coast, Forest, Highway, Industrial, InsideCity, Kitchen, LivingRoom, Mountain, Office, OpenCountry, Store, Street, Suburb and TallBuilding, with about 210 to 410 samples in each category.
Instead of using the original features, we employ the spatial pyramid features presented in \cite{6516503} for recognition, where each sample has 3000 dimensions.

\subsection{Experimental setup}
In our experiments, the nearest neighbor (1-NN) classifier is employed to classify samples using the Euclidean distance.
Parameters for methods are fine-tuned or set following authors' recommendations, except for neighbor size, which is uniformly set to the number of training samples in each class. 
For our method, we preset $s_{ij}=0$ if $x_i\notin N_k(x_j)$ where $k$ is neighbor size, then optimize the remaining $s_{ij}$.
For LPP, OLPP, RDR, LRLE, LRPP\_GRR and FSP the pre-constructed graph is weighted by 0-1 while others don't need to define the graph in advance.
To mitigate the effects of randomness, we implement all algorithms in the comparison experiments 10 times and report the mean recognition rates.
For each experiment, we randomly select a number of samples for each class as the training data (\#Tr), while using the remaining samples as the test data.
To improve the computational efficiency, we first apply PCA to reduce the dimensions of the data by preserving 98\% of the energy across all databases except dimension of 15-Scene is reduced to 198. 
Preserving 98\% of the energy means selecting only the first $k$ vectors corresponding to the $k$ largest eigenvalues that satisfy $\sum^k_{i=1}\lambda_i/\sum^n_{i=1}\lambda_i=0.98$, where $\lambda_i$ denotes the $i$-th largest eigenvalue of the covariance matrix of the original data and $n$ is the total number of eigenvalues.
These selected vectors are used to form the projection matrix of PCA for dimensionality reduction and then perform sample normalization e.g. $x_i=x_i/\|x_i\|_2$.
Before training the (1-NN) classifier, We also normalize both training and test samples to get recognition rate.

\subsection{Experiment result and analysis}
\begin{table*}
  \centering
  \caption{Mean Recognition Accuracies and Standard deviations (\%) of Various Algorithms on Six Different Databases}
  \resizebox{\linewidth}{!}{
  \begin{tabular}{|l|lcccccccccccc|}
    \hline
    Dataset & \#Tr & LPP & NPE & OLPP & PCAN & SOGFS & RJSE & RDR & LRLE & LRPP\_GRR & FSP & LRAGE & NGLGE \\
    \hline
    \multirow{4}{*}{EYaleB}
    & 10 & 50.02$\pm$1.90 & 57.86$\pm$1.02 & 43.07$\pm$1.27 & 83.19$\pm$0.97 & 61.74$\pm$1.83 & 45.76$\pm$1.27 & 54.05$\pm$1.65 & 50.54$\pm$1.28 & 85.45$\pm$0.93 & 81.20$\pm$1.32 & 39.39$\pm$1.99 & \textbf{85.91$\pm$0.93} \\
    & 15 & 58.75$\pm$2.29 & 64.44$\pm$1.67 & 52.55$\pm$0.81 & 87.35$\pm$0.51 & 68.83$\pm$4.22 & 54.34$\pm$0.88 & 59.44$\pm$2.25 & 56.04$\pm$2.05 & \textbf{90.27$\pm$0.62} & 85.56$\pm$0.40 & 53.02$\pm$4.19 & 89.64$\pm$0.57 \\
    & 20 & 64.85$\pm$1.63 & 68.96$\pm$0.61 & 58.72$\pm$0.82 & 89.14$\pm$0.92 & 75.62$\pm$0.92 & 60.21$\pm$1.00 & 63.31$\pm$1.32 & 60.60$\pm$1.27 & \textbf{91.44$\pm$0.59} & 86.63$\pm$1.11 & 61.14$\pm$3.67 & 91.22$\pm$0.80 \\
    & 25 & 67.96$\pm$1.62 & 72.64$\pm$1.26 & 62.23$\pm$1.10 & 89.99$\pm$0.89 & 78.33$\pm$0.86 & 63.48$\pm$0.94 & 65.48$\pm$0.99 & 63.31$\pm$0.95 & 92.11$\pm$0.65 & 88.29$\pm$0.96 & 67.81$\pm$3.59 & \textbf{92.38$\pm$0.41} \\
    \hline
    \multirow{4}{*}{YTC}
    & 10 & 70.79$\pm$0.89 & 71.98$\pm$0.99 & 76.04$\pm$1.03 & 66.86$\pm$0.86 & 72.77$\pm$1.07 & 76.18$\pm$1.05 & 76.09$\pm$0.95 & 75.67$\pm$0.98 & \textbf{76.99$\pm$1.08} & 72.39$\pm$0.92 & 75.30$\pm$1.06 & 76.27$\pm$1.03 \\
    & 15 & 79.29$\pm$0.96 & 79.84$\pm$0.80 & 82.37$\pm$0.86 & 75.60$\pm$0.83 & 79.31$\pm$0.92 & 82.43$\pm$0.79 & 82.50$\pm$0.78 & 81.84$\pm$0.82 & \textbf{82.54$\pm$0.75} & 79.26$\pm$1.22 & 81.17$\pm$0.92 & 82.50$\pm$0.81 \\
    & 20 & 83.14$\pm$0.67 & 84.16$\pm$0.75 & 85.48$\pm$0.59 & 80.19$\pm$0.45 & 82.83$\pm$0.55 & 85.49$\pm$0.60 & 85.39$\pm$0.59 & 84.91$\pm$0.62 & 84.88$\pm$0.51 & 82.63$\pm$0.81 & 84.36$\pm$0.90 & \textbf{85.59$\pm$0.61} \\
    & 25 & 85.82$\pm$0.48 & 86.83$\pm$0.41 & 87.75$\pm$0.54 & 83.24$\pm$0.60 & 85.44$\pm$0.67 & 87.75$\pm$0.56 & 87.67$\pm$0.52 & 87.12$\pm$0.47 & 86.62$\pm$0.53 & 84.81$\pm$0.51 & 86.65$\pm$0.65 & \textbf{87.83$\pm$0.54} \\
    \hline
    \multirow{3}{*}{Binalpha}
    & 10 & 38.31$\pm$1.52 & 39.25$\pm$1.56 & 60.60$\pm$1.67 & 17.13$\pm$1.17 & 57.99$\pm$1.91 & 60.21$\pm$1.87 & 60.30$\pm$1.77 & 60.11$\pm$1.85 & 49.12$\pm$4.32 & 24.54$\pm$2.85 & 60.58$\pm$1.75 & \textbf{61.84$\pm$1.49} \\
    & 15 & 54.27$\pm$1.56 & 55.54$\pm$1.81 & 64.69$\pm$1.46 & 25.57$\pm$1.59 & 62.37$\pm$1.10 & 64.48$\pm$1.05 & 64.73$\pm$1.16 & 64.09$\pm$1.47 & 54.54$\pm$2.25 & 27.09$\pm$3.65 & 64.02$\pm$1.65 & \textbf{65.50$\pm$0.97} \\
    & 20 & 60.53$\pm$2.32 & 61.86$\pm$1.48 & 67.06$\pm$1.29 & 30.91$\pm$2.02 & 62.68$\pm$2.48 & 66.61$\pm$1.29 & 66.65$\pm$1.00 & 67.19$\pm$1.42 & 55.92$\pm$3.45 & 26.71$\pm$1.69 & 66.56$\pm$1.67 & \textbf{68.11$\pm$0.98} \\
    \hline
    \multirow{4}{*}{USPS}
    & 10 & 77.63$\pm$1.74 & 79.22$\pm$1.69 & 81.96$\pm$1.04 & 74.18$\pm$2.21 & 81.46$\pm$2.08 & 84.01$\pm$0.95 & 82.32$\pm$1.14 & 81.16$\pm$1.26 & 79.58$\pm$1.62 & 68.12$\pm$2.97 & 82.99$\pm$1.40 & \textbf{84.05$\pm$0.97} \\
    & 20 & 85.61$\pm$0.61 & 86.12$\pm$0.76 & 86.40$\pm$0.77 & 83.28$\pm$1.55 & 86.44$\pm$0.91 & 87.75$\pm$0.67 & 86.16$\pm$0.80 & 86.55$\pm$0.76 & 84.12$\pm$1.06 & 76.30$\pm$2.31 & 86.96$\pm$1.13 & \textbf{87.81$\pm$0.60} \\
    & 30 & 88.55$\pm$0.60 & 89.21$\pm$0.41 & 88.94$\pm$0.50 & 86.31$\pm$0.91 & 89.58$\pm$0.61 & 90.23$\pm$0.29 & 88.68$\pm$0.55 & 89.12$\pm$0.54 & 87.52$\pm$0.64 & 82.70$\pm$1.06 & 89.57$\pm$0.41 & \textbf{90.24$\pm$0.26} \\
    & 40 & 90.29$\pm$0.41 & 90.78$\pm$0.45 & 90.36$\pm$0.53 & 87.69$\pm$0.56 & 90.88$\pm$0.99 & 91.23$\pm$0.40 & 90.25$\pm$0.41 & 90.58$\pm$0.32 & 89.69$\pm$0.60 & 84.31$\pm$1.18 & 90.71$\pm$0.35 & \textbf{91.28$\pm$0.40} \\
    \hline
    \multirow{4}{*}{ETH80}
    & 10 & 49.11$\pm$1.42 & 49.65$\pm$1.73 & 58.27$\pm$1.19 & 58.56$\pm$1.20 & 36.77$\pm$2.35 & 57.97$\pm$1.72 & 59.13$\pm$1.55 & \textbf{59.47$\pm$1.56} & 51.35$\pm$3.76 & 46.65$\pm$1.89 & 58.53$\pm$1.34 & 58.71$\pm$1.24 \\
    & 20 & 29.20$\pm$3.56 & 28.32$\pm$3.21 & 53.84$\pm$1.38 & 64.68$\pm$1.28 & 40.59$\pm$1.76 & 63.69$\pm$1.43 & 64.58$\pm$0.82 & 64.59$\pm$0.69 & 58.80$\pm$2.59 & 30.21$\pm$1.17 & 63.07$\pm$1.55 & \textbf{65.19$\pm$1.16} \\
    & 30 & 57.86$\pm$1.17 & 62.49$\pm$0.71 & 61.14$\pm$1.06 & 68.40$\pm$0.69 & 46.46$\pm$1.06 & 67.40$\pm$0.58 & 67.32$\pm$1.05 & 67.07$\pm$1.11 & 62.79$\pm$1.30 & 36.95$\pm$1.32 & 66.60$\pm$0.59 & \textbf{68.98$\pm$0.95} \\
    & 40 & 65.94$\pm$1.13 & 68.20$\pm$1.13 & 66.28$\pm$1.08 & 70.94$\pm$0.99 & 51.51$\pm$3.12 & 70.09$\pm$1.04 & 70.54$\pm$1.12 & 69.75$\pm$1.18 & 67.07$\pm$1.14 & 40.71$\pm$0.97 & 68.78$\pm$1.02 & \textbf{71.66$\pm$1.19} \\
    \hline
    \multirow{4}{*}{15-Scene}
    & 10 & 80.96$\pm$1.89 & 82.49$\pm$1.31 & 89.14$\pm$0.86 & 88.74$\pm$1.01 & 84.85$\pm$2.26 & 89.67$\pm$0.92 & 89.68$\pm$0.85 & 88.46$\pm$1.09 & 86.38$\pm$0.85 & 86.86$\pm$1.31 & 89.40$\pm$0.85 & \textbf{90.13$\pm$1.02} \\
    & 20 & 90.05$\pm$1.16 & 87.20$\pm$1.04 & 93.05$\pm$0.71 & 92.60$\pm$0.67 & 89.57$\pm$1.47 & 92.76$\pm$0.73 & 93.36$\pm$0.69 & 92.43$\pm$0.75 & 90.69$\pm$1.17 & 77.45$\pm$0.99 & 93.32$\pm$0.75 & \textbf{93.91$\pm$0.74} \\
    & 30 & 94.07$\pm$0.64 & 90.86$\pm$0.65 & 94.98$\pm$0.52 & 94.06$\pm$0.44 & 92.39$\pm$0.62 & 94.52$\pm$0.45 & 94.97$\pm$0.46 & 94.36$\pm$0.56 & 93.01$\pm$0.61 & 81.89$\pm$1.26 & 95.08$\pm$0.41 & \textbf{95.59$\pm$0.39} \\
    & 40 & 95.00$\pm$0.32 & 92.06$\pm$0.32 & 95.52$\pm$0.35 & 94.60$\pm$0.29 & 93.21$\pm$0.50 & 95.00$\pm$0.26 & 95.55$\pm$0.33 & 95.14$\pm$0.31 & 93.80$\pm$0.52 & 85.63$\pm$0.97 & 95.68$\pm$0.33 & \textbf{96.06$\pm$0.29} \\
    \hline
  \end{tabular}
  }
  \label{tabel}
\end{table*}

The results of each algorithm across diverse datasets are summarized in Table \ref{tabel} leading to the following conclusions:

1) LRPP\_GRR and our method excel in classification accuracy on EYaleB and YTC datasets, surpassing other methods. it implies that combining the graph regularization for local structure and low-rank representation for global structure proves to be beneficial.

2) Our method outperforms LRPP\_GRR and others on Binalpha and USPS datasets, which may implies that neighborhood-adaptive graph learning and $\ell_{2,0}$-norm feature selection show effectiveness in reducing inter-class margins and filtering redundant features. 

3) Form the experiment results of ETH80 dataset, our method competes well, particularly with 20, 30, or 40 training samples per class. LRLE excels with 10 samples per class, indicating our method's ability to leverage discriminative information with ample training samples.

4) On the 15-Scene dataset, our method consistently outperforms others. Notably, achieving over 90\% accuracy with a small training set demonstrates strong generalization.

\subsection{Parameters sensitivity study}
\begin{figure}
  \centering
  \subfigure[EYaleB]{
    \includegraphics[width=0.29\linewidth]{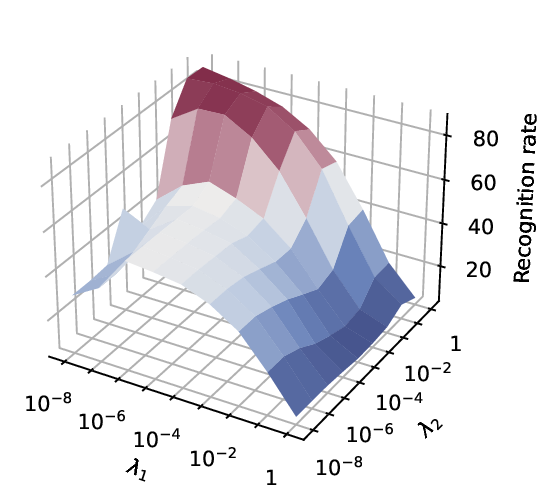}
  }
  \subfigure[YTC]{
    \includegraphics[width=0.29\linewidth]{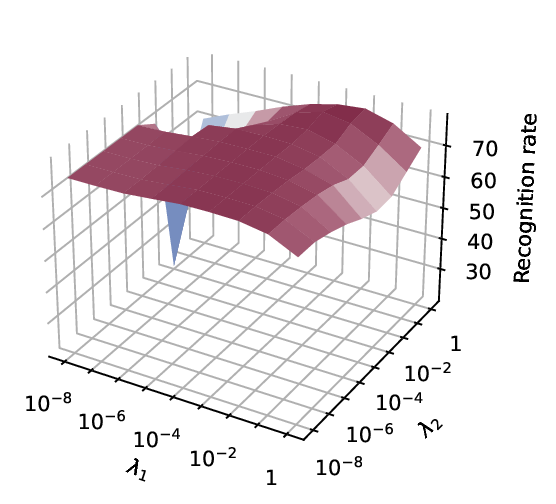}
  }
  \subfigure[Binalpha]{
    \includegraphics[width=0.29\linewidth]{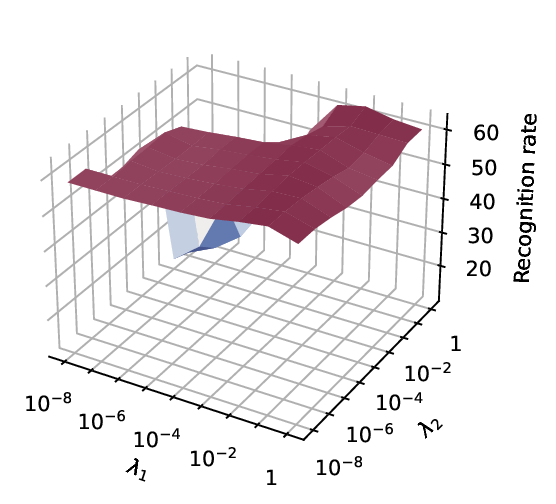}
  }
  
  \subfigure[USPS]{
    \includegraphics[width=0.29\linewidth]{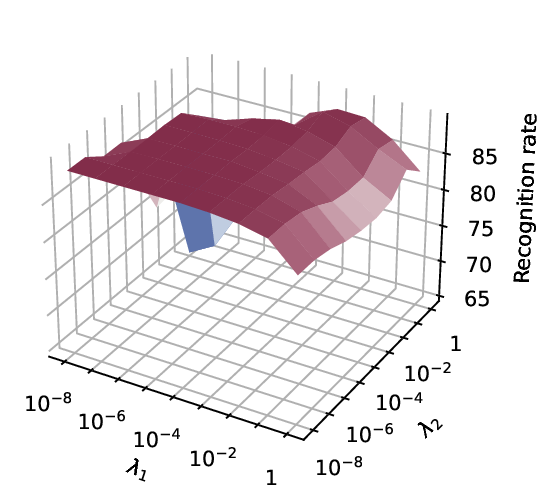}
  }
  \subfigure[ETH80]{
    \includegraphics[width=0.29\linewidth]{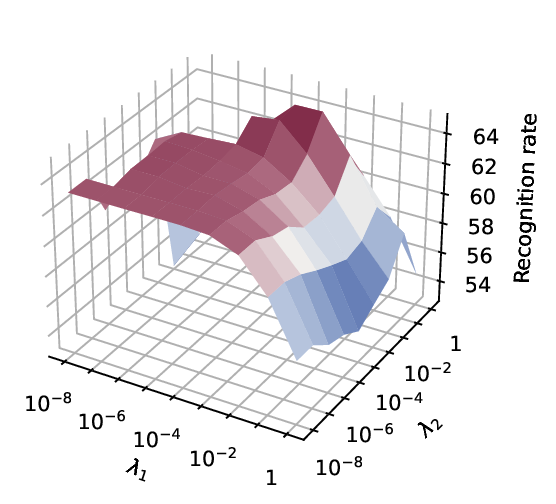}
  }
  \subfigure[15-Scene]{
    \includegraphics[width=0.29\linewidth]{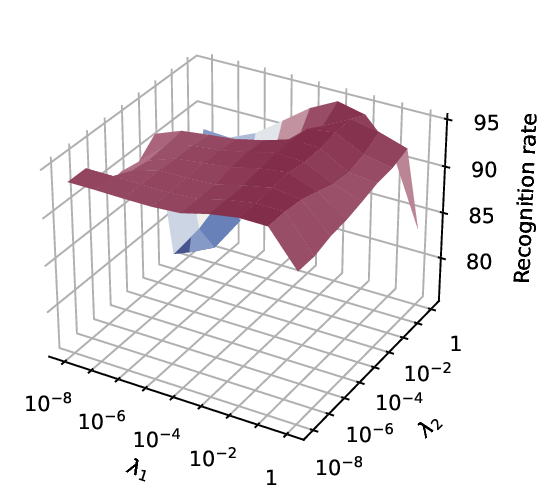}
  }
  \caption{Recognition rate versus the parameters $\lambda_1$ and $\lambda_2$ on six databases with training samples of 10, 10, 10, 20, 20 and 20 per category respectively.}
  \label{fig:lambda12}
\end{figure}
\begin{figure}
  \centering
   \includegraphics[width=0.9\linewidth]{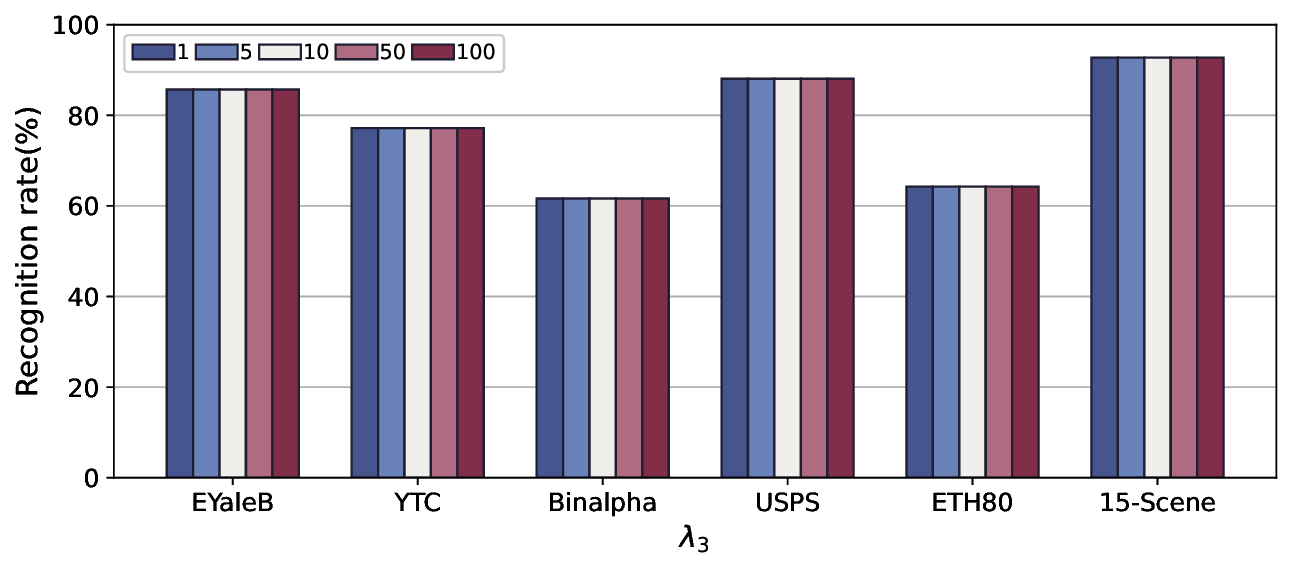}
   \caption{Classification performance (\%) versus hyper-paramater $\lambda_3$ on six different databases with 10, 10, 10, 20, 20 and 20 training samples per class, respectively.}
   \label{fig:lambda3}
\end{figure}
The classification performance of model (\ref{model}) can be influenced by four tunable parameters, namely $\lambda_1$, $\lambda_2$, $\lambda_3$, and $\alpha$.
However, adaptively selecting suitable parameters remains an open problem for various tasks, and no efficient solution has been proposed thus far.
To simplify the process of discovering the optimum parameters, we fix $\alpha$ to $\mathrm{max}(m, \lfloor0.9d\rfloor)$, e.g., selecting 90\% original feature.
In our experiments, we observed minimal relevance of $\lambda_3$ to other parameters.
As a result, we determine $\lambda_3$ values independently.
However, $\lambda_1$ and $\lambda_2$ exhibit interdependence, leading us to perform a grid search to identify optimal values for these two parameters.
Specifically, $\lambda_1$ and $\lambda_2$ are selected from $\{10^{-8}, 10^{-7}, 10^{-6}, 10^{-5}, 10^{-4}, 10^{-3}, 10^{-2}, 0.1, 1\}$ and $\lambda_3$ is chosen from $\{1, 5, 10, 50, 100\}$.
Fig.~\ref{fig:lambda12} illustrates the performance of the proposed algorithm under different parameters combining across six databases.
By fixing other hyper-parameters, we show the impact of different $\lambda_3$ on classification performance in Fig.~\ref{fig:lambda3}, which verifies that our model is insensitive to the selection of $\lambda_3$ in its candidate set.

\subsection{Research on dimensionality sensitivity}
\begin{figure}
  \centering
  \subfigure[EYaleB]{
    \includegraphics[width=0.29\linewidth]{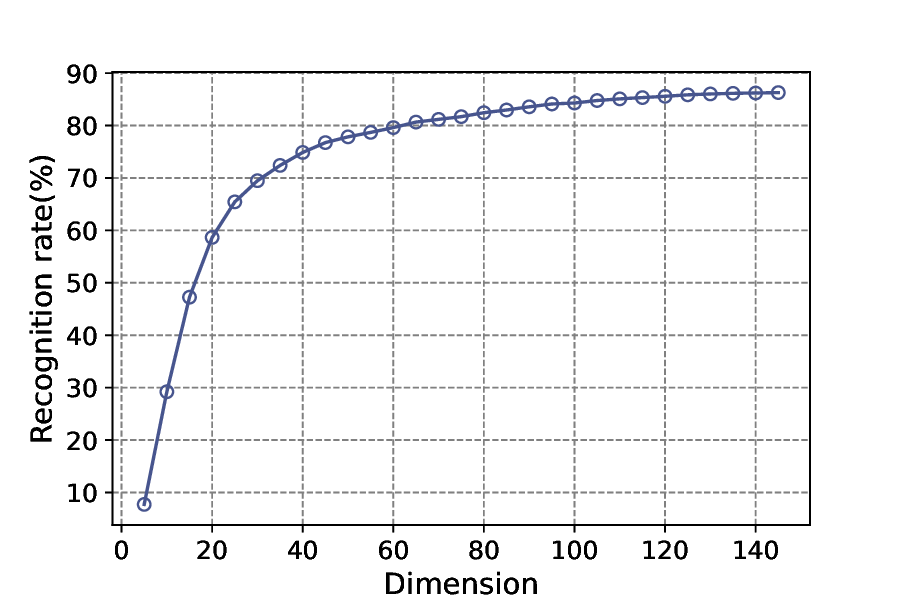}
  }
  \subfigure[YTC]{
    \includegraphics[width=0.29\linewidth]{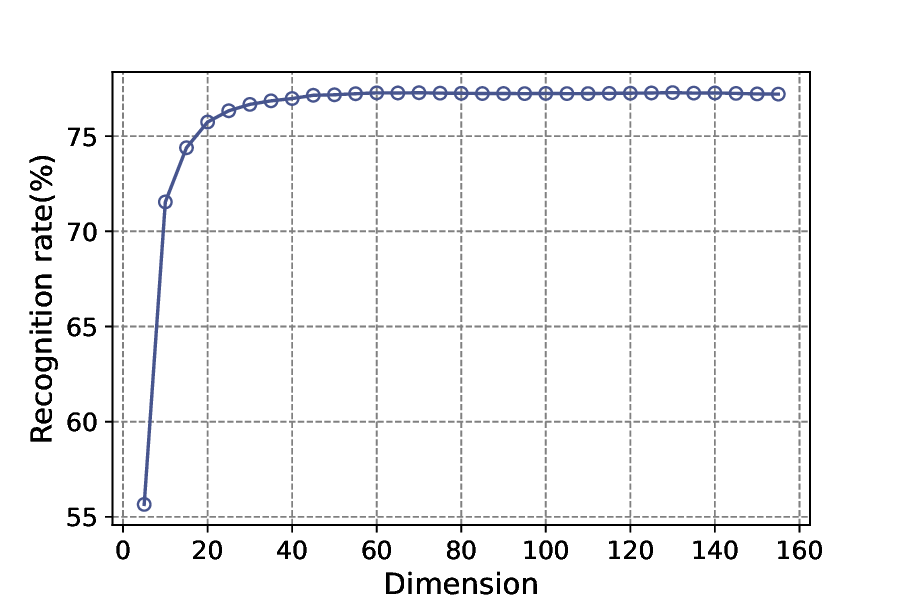}
  }
  \subfigure[Binalpha]{
    \includegraphics[width=0.29\linewidth]{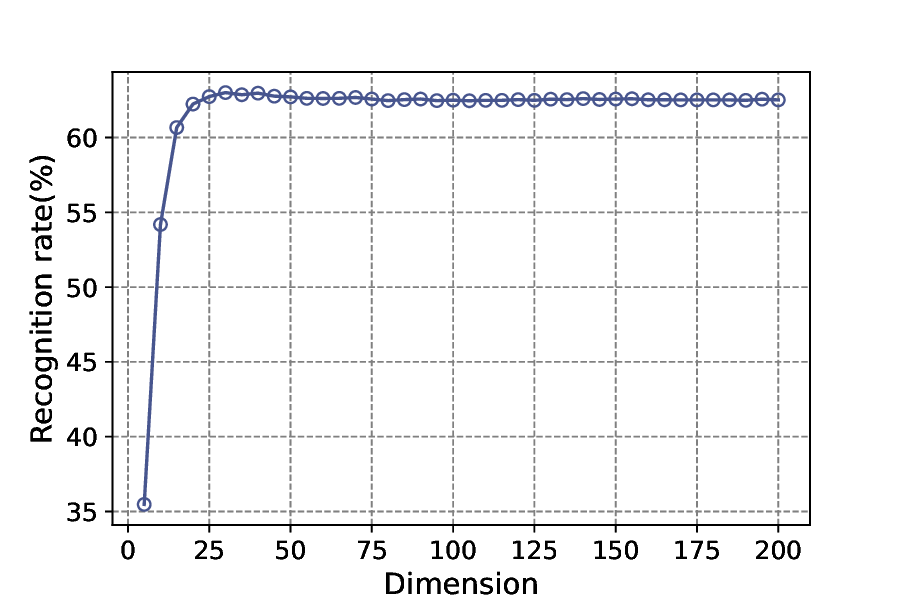}
  }
  
  \subfigure[USPS]{
    \includegraphics[width=0.29\linewidth]{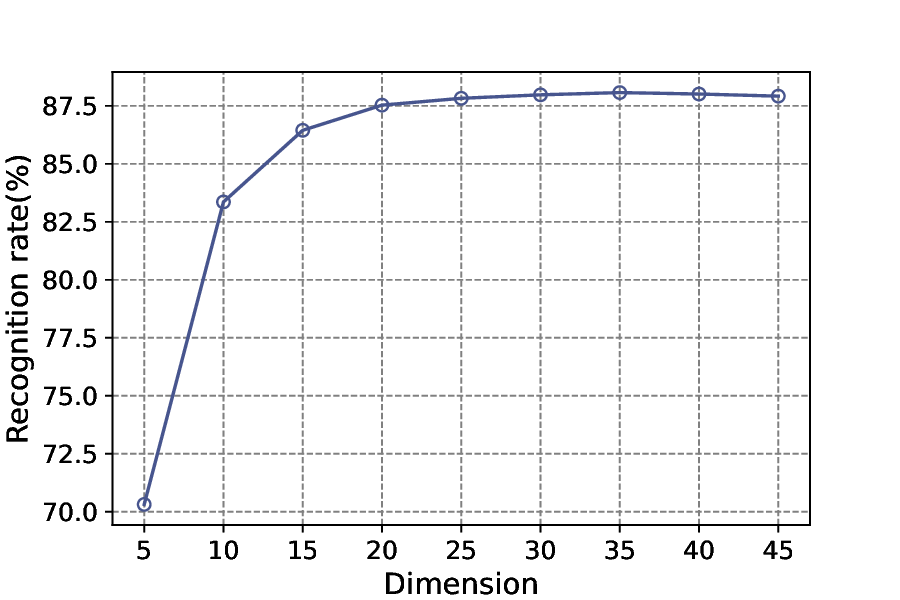}
  }
  \subfigure[ETH80]{
    \includegraphics[width=0.29\linewidth]{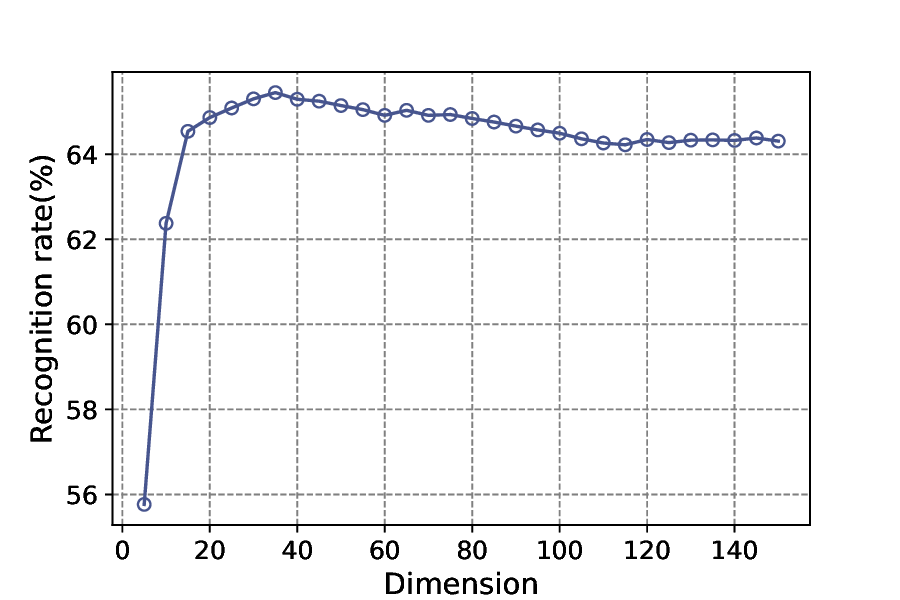}
  }
  \subfigure[15-Scene]{
    \includegraphics[width=0.29\linewidth]{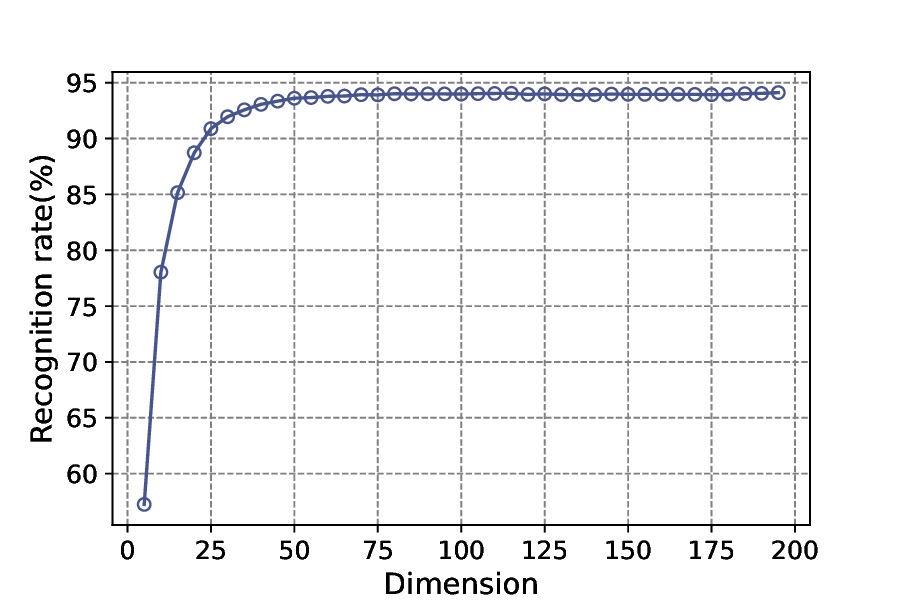}
  }
  \caption{Recognition rate versus the reduced dimension on six databases with training samples of 10, 10, 10, 20, 20 and 20 per category respectively.}
  \label{fig:dim}
\end{figure}
To conduct feature extraction, we assessed the proposed method at various dimensionalities on six distinct databases respectively.
Fig.~\ref{fig:dim} depicts the recognition rate versus reduced dimension.
From the figure, we can observe that the model delivers satisfactory performance across a broad range of feature dimensions, provided that the dimensionality is not excessively low.
In our reported results, we employed reduced dimensions of 140, 150, 200, 40, 70 and 140 for all methods on EYaleB, YTC, Binalpha, USPS, ETH80 and 15-Scene databases, respectively.

\subsection{Convergence study}
\begin{figure}
  \centering
  \subfigure[EYaleB]{
    \includegraphics[width=0.282\linewidth]{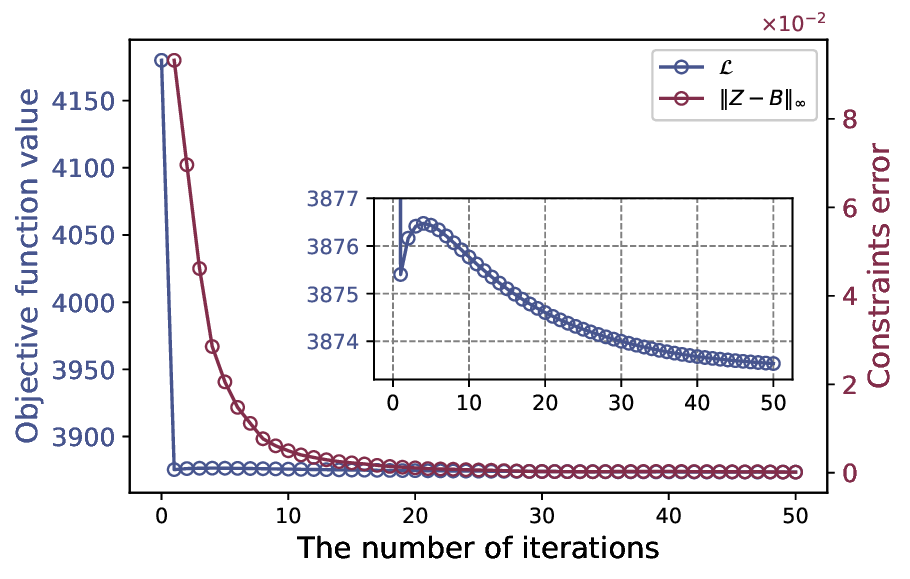}
  }
  \subfigure[YTC]{
    \includegraphics[width=0.289\linewidth]{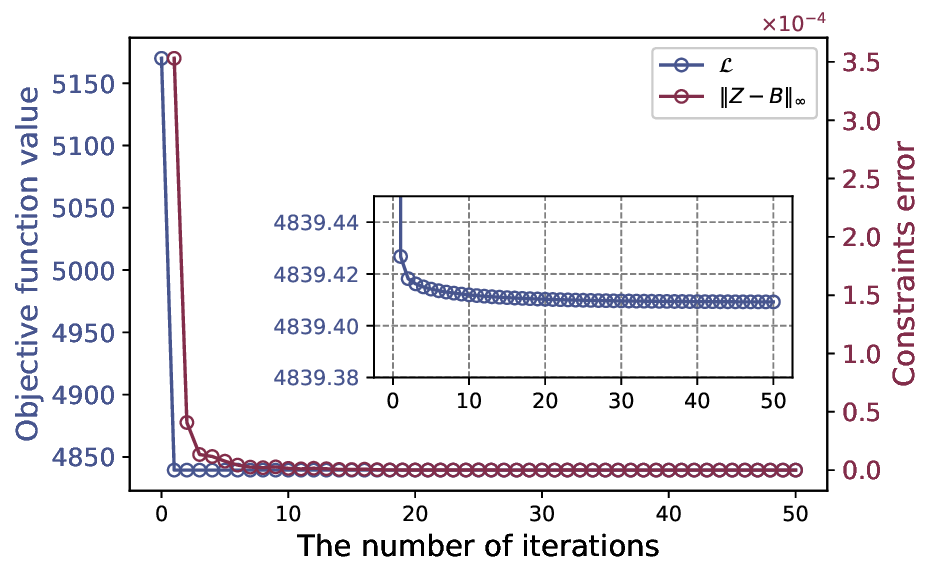}
  }
  \subfigure[Binalpha]{
    \includegraphics[width=0.29\linewidth]{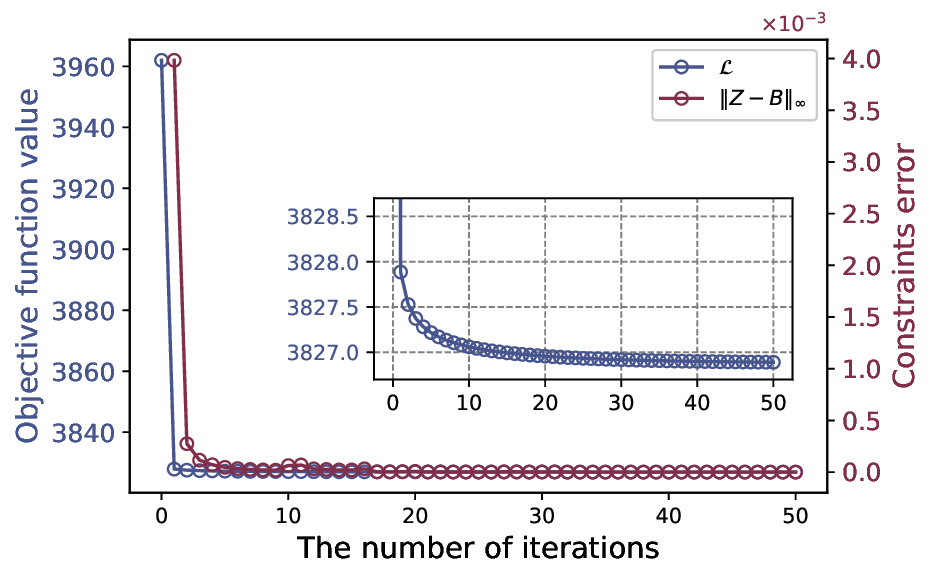}
  }
  
  \subfigure[USPS]{
    \includegraphics[width=0.289\linewidth]{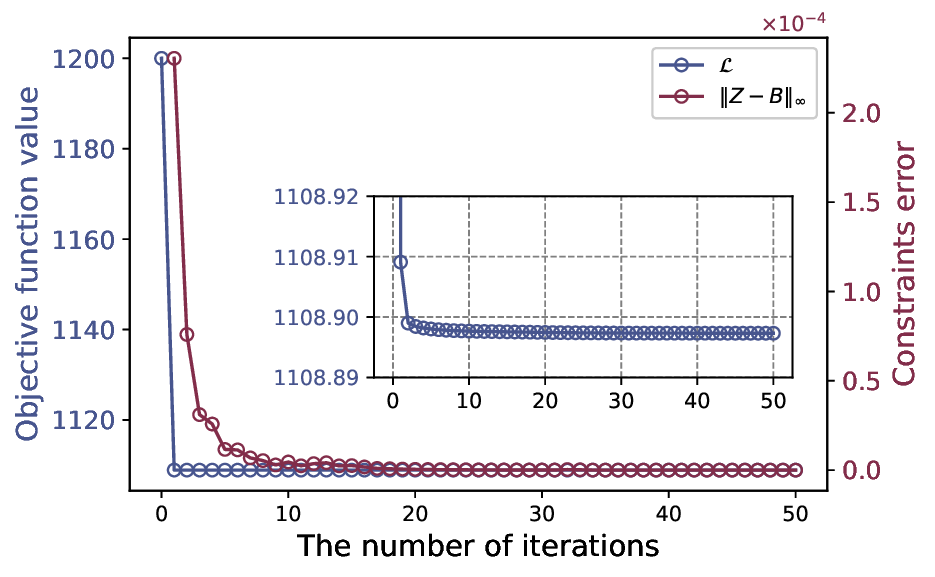}
  }
  \subfigure[ETH80]{
    \includegraphics[width=0.285\linewidth]{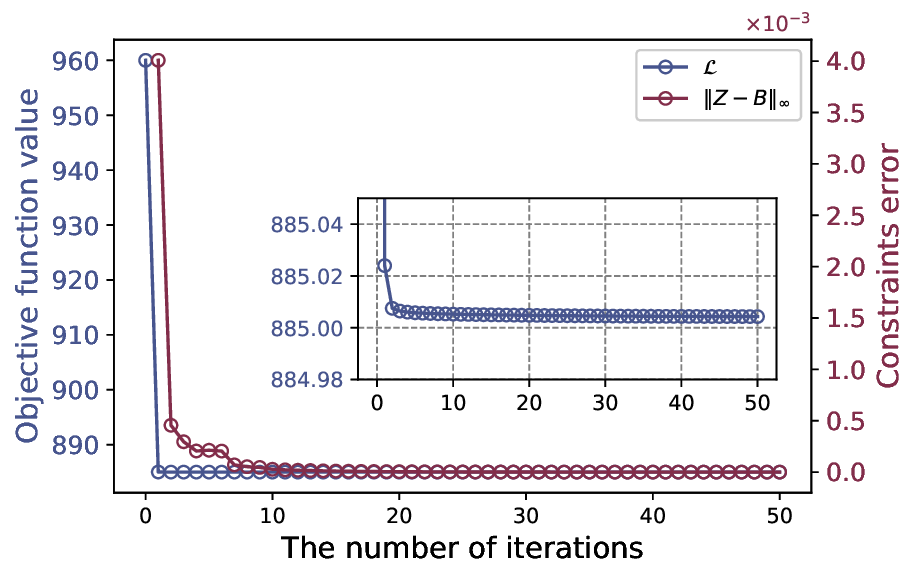}
  }
  \subfigure[15-Scene]{
    \includegraphics[width=0.29\linewidth]{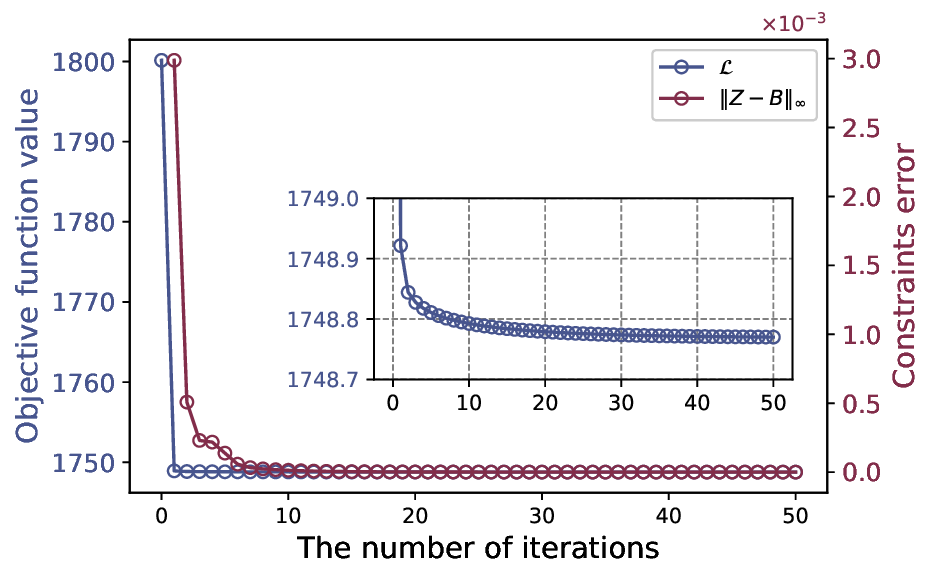}
  }
  \caption{Convergence curves of our method on six databases with training samples of 10, 10, 10, 20, 20 and 20 per category respectively.}
  \label{fig:converg}
\end{figure}
In this section, we establish the convergence property of our approach from an experimental perspective, conducting six experiments on the EYaleB, YTC, Binalpha, USPS, ETH80 and 15-Scene databases.
Fig.~\ref{fig:converg} shows the objective value and constraint error versus different iterations for the six databases.
The plots demonstrate that the objective loss and constraints error of our learning model (\ref{model}) steadily decrease and eventually stabilize, indicating that the exploited optimization algorithm exhibits favorable convergence properties.

\section{Conclusion}\label{sec:con}
In summary, this paper introduces a generalized linear graph embedding method. 
Firstly, we propose a neighborhood-adaptive graph learning approach that does not require predefining the neighborhood size. 
Secondly, the diversity in mining latent patterns is determined by the low-rank representation and the $\ell_{2,0}$ column sparsity of the projection matrix. 
Experimental results demonstrate the outstanding performance of the proposed method in handling various scenarios.

{
\bibliographystyle{IEEEtran}
\bibliography{IEEEabrv,egbib}
}

\end{document}